\documentclass[letterpaper, 10pt, conference]{ieeeconf}

\IEEEoverridecommandlockouts
\usepackage{amsmath,amsfonts}
\usepackage{array}
\usepackage[font=footnotesize]{subcaption}
\usepackage{textcomp}
\usepackage{stfloats}
\usepackage{url}
\usepackage{verbatim}
\usepackage{graphicx}
\graphicspath{{fig/}}
\usepackage{cite}
\usepackage{xcolor}
\usepackage{multirow}
\usepackage{booktabs}
\title{\LARGE \bf
Safe and Efficient Navigation Considering Texting Pedestrians
via Attention-Aware Cost-map Tuning
}

\author{Shunya Tadano, Yusuke Tamura,
Ankit A. Ravankar, and Yasuhisa Hirata%
\thanks{The authors are with the Department of Robotics, Graduate School of Engineering, Tohoku University, 6-6-01 Aramaki Aza Aoba, Aoba-ku, Sendai, Japan,
{\tt\small tadano.shunya.r4@dc.tohoku.ac.jp}}%
}

\begin{document}

\maketitle
\thispagestyle{empty}
\pagestyle{empty}

\begin{abstract}

Smartphone-distracted pedestrians deviate from social norms unpredictably and cannot be relied upon to yield to approaching
robots --- a safety challenge that existing navigation methods are poorly equipped to handle, since they treat all pedestrians
uniformly regardless of attention state. We propose an  attention-aware navigation framework that classifies each
pedestrian as attentive or distracted in real-time using 3D  skeletal features and a Random Forest classifier, then dynamically
widens cost-map safety margins around distracted individuals to  drive proactive avoidance. A within-subjects user study with 12
participants across 252 trials demonstrates significant  improvements in perceived smoothness ($p=0.003$) and  predictability ($p=0.034$) over both a LiDAR-only baseline and a
trajectory-prediction-only method, without any increase in travel  time. Notably, trajectory prediction alone does not improve
subjective navigation quality unless coupled with  attention-state-aware cost adjustment --- confirming that  appropriately responding to \textit{who} is in front of the robot
can matter more than trajectory smoothness alone.

\end{abstract}

%\begin{IEEEkeywords}
%    Social Navigation, Human-robot Interaction, Attention-aware Navigation, Real-time Classification, Motion Planning.
%\end{IEEEkeywords}

%%%%%%%%%%%%%%%%%%%%%%%%%%%%%%%%%%%%%%%%%%%%%%%%%%%%%%%%%%%%%%%%%%%%%%%%%%%%%%%%

\section{Introduction}

Autonomous robots operating in shared human spaces must do more than avoid collisions --- they must navigate in ways that are
safe, predictable, and socially acceptable to the people around them~\cite{francis2025principles}. This requires robots to continuously interpret pedestrian behavior and adapt their motion accordingly~\cite{mavrogiannis2023core}.
Service robots deployed in hospitals, offices, and care facilities face this challenge routinely, where the cost of getting it wrong
is not merely inefficiency but physical risk. Despite significant advances in social navigation, however, a persistent gap remains:
existing methods treat all pedestrians as attentive, cooperative agents --- an assumption that breaks down the moment someone picks
up their phone.
This gap is most consequential for service robots in healthcare and care facilities, which routinely share narrow corridors with distracted patients, visitors, and staff~\cite{morgan2022robots}.

Smartphone-distracted pedestrians present a qualitatively  different navigation challenge. Observational studies report that
12\% to 45\% of pedestrians in public spaces are actively distracted while walking, with texting shown to moderately
increase the rate of collisions and near-misses~\cite{simmons2020plight}. Distraction induces ``inattentional blindness''~\cite{hyman2010did},
alters gait parameters including walking speed and stride length~\cite{sajewicz2023texting}, and significantly slows
responses to environmental changes~\cite{murakami2021mutual}. Crucially, a distracted pedestrian is unlikely to notice an
approaching robot or adjust their path --- placing the full burden of conflict resolution on the robot itself. No existing navigation
approach accounts for this asymmetry.

\begin{figure}[tb]
    \centering
    \includegraphics[width=3.4in]{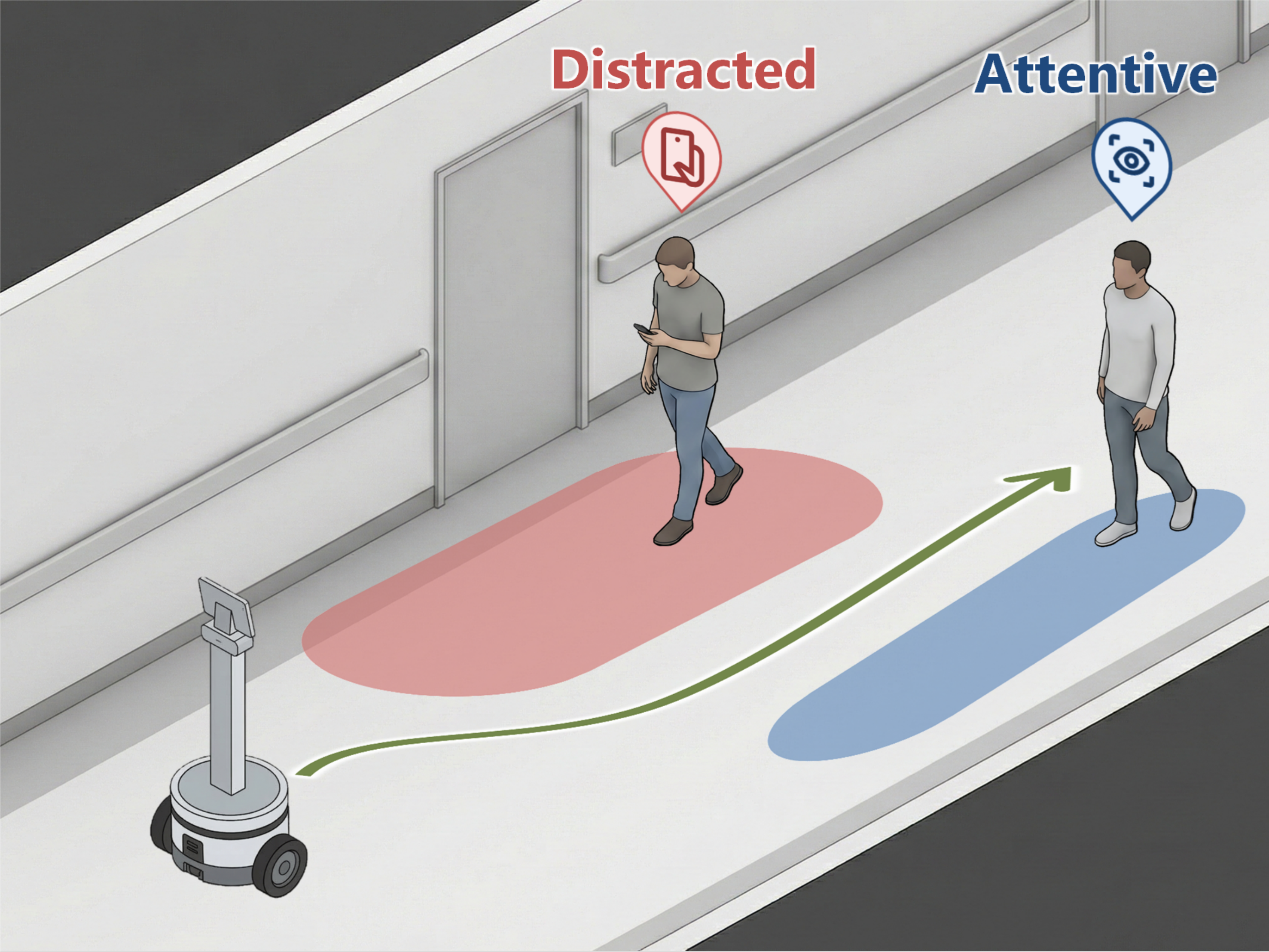}
    \caption{An overview of the proposed attention-aware cost-map tuning. A robot secures a standard safety margin for an attentive pedestrian (right) and a larger margin for a smartphone-distracted pedestrian (left).}
    \label{fig:concept}
\end{figure}

This paper proposes an attention-aware navigation framework that directly addresses this gap. The system classifies each pedestrian as attentive or distracted in real-time from 3D skeletal information and uses this classification to dynamically adjust cost-map safety margins during motion planning. As shown in Fig.~\ref{fig:concept}, wider avoidance regions are assigned
around distracted pedestrians, allowing the robot to plan proactively rather than react to an unfolding conflict. A within-subjects experiment with 12 participants across 252 trials demonstrates that this approach significantly improves perceived smoothness and predictability of robot navigation while maintaining travel efficiency.

The contributions of this research are as follows:

\begin{itemize}
    \setlength{\itemsep}{2pt}
    \setlength{\topsep}{3pt}
    \setlength{\parskip}{0pt}

    \item \textbf{Skeletal-based Distraction Classification:} A pose-based module that classifies pedestrians as attentive or
    distracted in real-time from 3D skeletal keypoints using a lightweight Random Forest classifier, designed for deployment
    on mobile robot hardware.
    \item \textbf{Attention-Aware Cost Tuning for Legible Avoidance:} A dynamic cost-map layer that modulates safety margins based on classified attention state, making avoidance behavior more legible and anticipatable around non-cooperative pedestrians.
    \item \textbf{Empirical Validation:} A user study
    demonstrating that attention-state classification --- not trajectory prediction alone --- is the critical factor connecting objective motion smoothness to perceived navigation
    quality.
\end{itemize}

\begin{figure*}[htbp]
    \centering
    \includegraphics[width=6.98in]{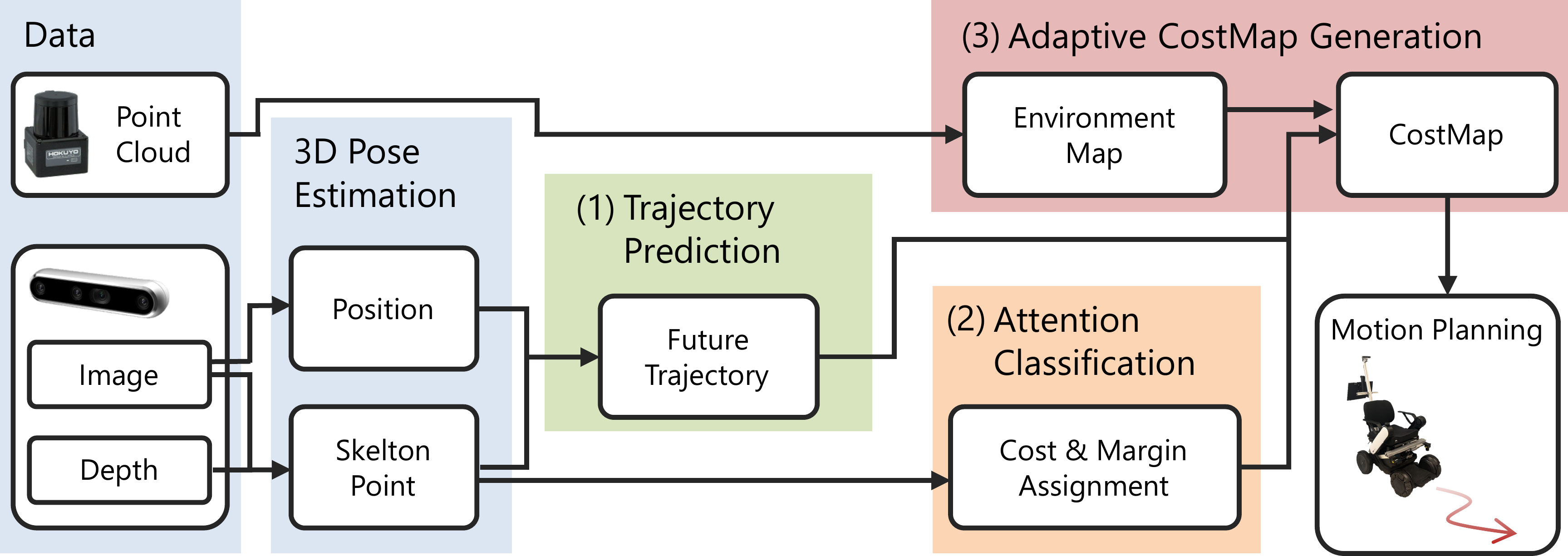}
    \caption{Framework of the proposed system. The system consists of three main modules: (1) pedestrian trajectory prediction based on historical position data, (2) Attention Classification using 3D skeletal keypoint extraction (33 points) from RGB-D camera images, (3) adaptive CostMap Generation using predicted trajectories based on attention state classification results.}
    \label{fig:framework}
\end{figure*}

\section{Related Works}

\subsection{Trajectory Prediction with Uncertainty Modeling}

Modeling uncertainty in pedestrian trajectory prediction is central to safe robot navigation. Deep learning approaches such
as Social-LSTM~\cite{alahi2016social} and Social-GAN~\cite{gupta2018social} capture this uncertainty by
generating multiple plausible future trajectories from observed motion history. More structured approaches model specific
pedestrian intentions and behaviors --- such as whether a pedestrian will yield at an intersection --- to improve both accuracy and
interpretability~\cite{tamura2012development,tanno2023trajectory}. Trajectron++ takes this further by producing full predictive distributions
with per-mode confidence estimates~\cite{salzmann2020trajectron++}. Despite these advances, a common limitation persists: all pedestrians are assigned identical uncertainty models, regardless of attention state. A distracted pedestrian is less predictable than an attentive one, and current methods do not
reflect this distinction.

\subsection{Pedestrian Attention State Recognition}
Recognizing pedestrian attention states is important for navigation safety, since a pedestrian's behavioral
predictability is closely tied to their level of environmental awareness. Gaze-based approaches estimate
attention targets from eye direction~\cite{tu2025real} or head orientation~\cite{benfold2009guiding}, and can achieve
high accuracy when conditions permit. However, they require high-resolution facial imagery and wide-angle cameras, making
them computationally expensive and difficult to deploy on  mobile robot hardware.

Body pose offers a more practical alternative. Smartphone-distracted pedestrians adopt characteristic postures --- raised forearms, downward head tilt --- that
are detectable from whole-body skeletal information. Detection of texting while walking has been demonstrated
using image-based methods~\cite{terao2025detection} and 3D LiDAR~\cite{wu2020smartphone}, both of which can operate at
useful ranges without requiring facial detail. Deep learning classifiers such as CNNs and Vision Transformers
achieve strong action recognition performance~\cite{feichtenhofer2020x3d}, but their
computational demands can introduce unacceptable latency on resource-constrained robot platforms. Lightweight classifiers
with interpretable feature representations are therefore better suited for real-time deployment in navigation loops.
Critically, however, existing work in this area focuses on classification accuracy in isolation --- the problem of
integrating attention state recognition into live robot navigation has not been addressed.

\subsection{Adaptive Cost Mapping in Social Navigation}

Traditional reactive planners such as the Social Force Model~\cite{helbing1995social} represent crowds as
homogeneous particles governed by repulsive potentials. This works reasonably well for attentive pedestrians who
respond to the robot's presence, but fails entirely for distracted individuals who do not perceive or react to
those forces. Cost-map-based planners offer a more flexible alternative by encoding social constraints
directly into the planning objective. Algorithms such as Dynamic Window Approach (DWA)~\cite{fox2002dynamic} and Time Elastic Band (TEB)~\cite{rosmann2012trajectory}
are cost-aware and can incorporate social constraints beyond binary obstacle information --- for example,
modeling human personal space as a soft cost to produce more socially acceptable trajectories~\cite{bilen2024social,fernandez2019towards,ravankar2020safe}.
Related studies target smoothness and comfort directly — respecting pedestrian personal space in indoor coexistence~\cite{tamura2010smooth}, and preserving smooth, collision-free trajectories around dynamic obstacles in semi-structured outdoor environments~\cite{ravankar2021autonomous}.
Virtual obstacle representations offer a complementary mechanism, injecting risk regions into the planner without modifying the underlying map~\cite{ravankar2019virtual}.
TEB-style optimization has been developed to efficiently handle dynamic constraints in real time~\cite{rosmann2013efficient}. 

Adaptive cost adjustment has been applied to a range of social contexts, including emotional state~\cite{bilen2024social}
and conversational groups~\cite{luo2024gson}. These approaches show that dynamic cost modulation can meaningfully improve social navigation behavior. However, they operate at the level of group dynamics or broad
affective states, and do not consider individual pedestrian attention as a factor in real-time risk assessment.

Taken together, the three areas reviewed above each address part of the problem but none addresses it
completely. Trajectory prediction methods do not distinguish between attentive and distracted pedestrians.
Attention recognition methods are not integrated into navigation pipelines. Cost-map approaches adapt to social
context but overlook the individual attention state as a driver of collision risk. This work bridges these gaps by
combining real-time attention classification with adaptive cost-map tuning, producing navigation behavior that is
directly responsive to whether a pedestrian can be expected to cooperate in avoiding a conflict.

\section{Methodology}
The proposed system addresses three interconnected limitations of existing approaches: uniform uncertainty modeling across
all pedestrians, attention recognition that is evaluated in isolation from navigation, and cost-map adaptation that
overlooks individual attention state as a risk factor. As shown in Fig.~\ref{fig:framework}, the system integrates
three modules --- real-time attention classification, pedestrian trajectory prediction using the Human Scene
Transformer (HST)~\cite{salzmann2023robots}, and adaptive cost-map generation --- to produce navigation behavior that
is directly responsive to each pedestrian's attention state. Point clouds, RGB images, and depth images are acquired from
robot-mounted RGB-D cameras and LiDAR. From these inputs, skeletal estimation models generate pedestrian positions and
3D skeleton representations, which feed into both the trajectory prediction and attention classification modules.
Predicted trajectory distributions are then incorporated into the cost map, and safe paths are generated through
motion planning using TEB~\cite{rosmann2012trajectory}.

\subsection{Attention Classification}

To detect pedestrian attention states, 3D skeletal information is extracted from RGB images using MediaPipe
BlazePose~\cite{bazarevsky2020blazepose}, which yields 33 keypoints $P = \{p_1, ..., p_{33}\}$, where each keypoint
$p_i = (x_i, y_i, z_i)$ is represented in camera coordinates. To ensure invariance to pedestrian position
and scale while preserving the relative orientation toward the robot, the keypoints are normalized by centering the
coordinate system at the midpoint of the hip joints ($p_{23}$ and $p_{24}$) and scaling all keypoints by the
maximum distance from this origin. This preserves the pedestrian's view-dependent posture, allowing the classifier
to implicitly learn the approach angle relative to the robot. The resulting 99-dimensional feature vector is
defined as $f = [p'_1, ..., p'_{33}]$:

\begin{equation}
    c_{hip} = \frac{p_{23} + p_{24}}{2}, \quad p'_i =
    \frac{p_i - c_{hip}}{\max_j \| p_j - c_{hip} \|}
\end{equation}

Real-time classification on mobile robot hardware demands low computational cost without sacrificing reliability. To
meet these constraints, we employ a Random Forest classifier with 300 decision trees --- sufficient for robust predictive
performance while remaining lightweight enough for on-board deployment. A maximum tree depth of 20 is used to balance
model expressiveness against overfitting to the training data. Classifier performance is validated in Section~V-A.

\begin{figure}[tb]
    \centering
    \includegraphics[width=\linewidth]{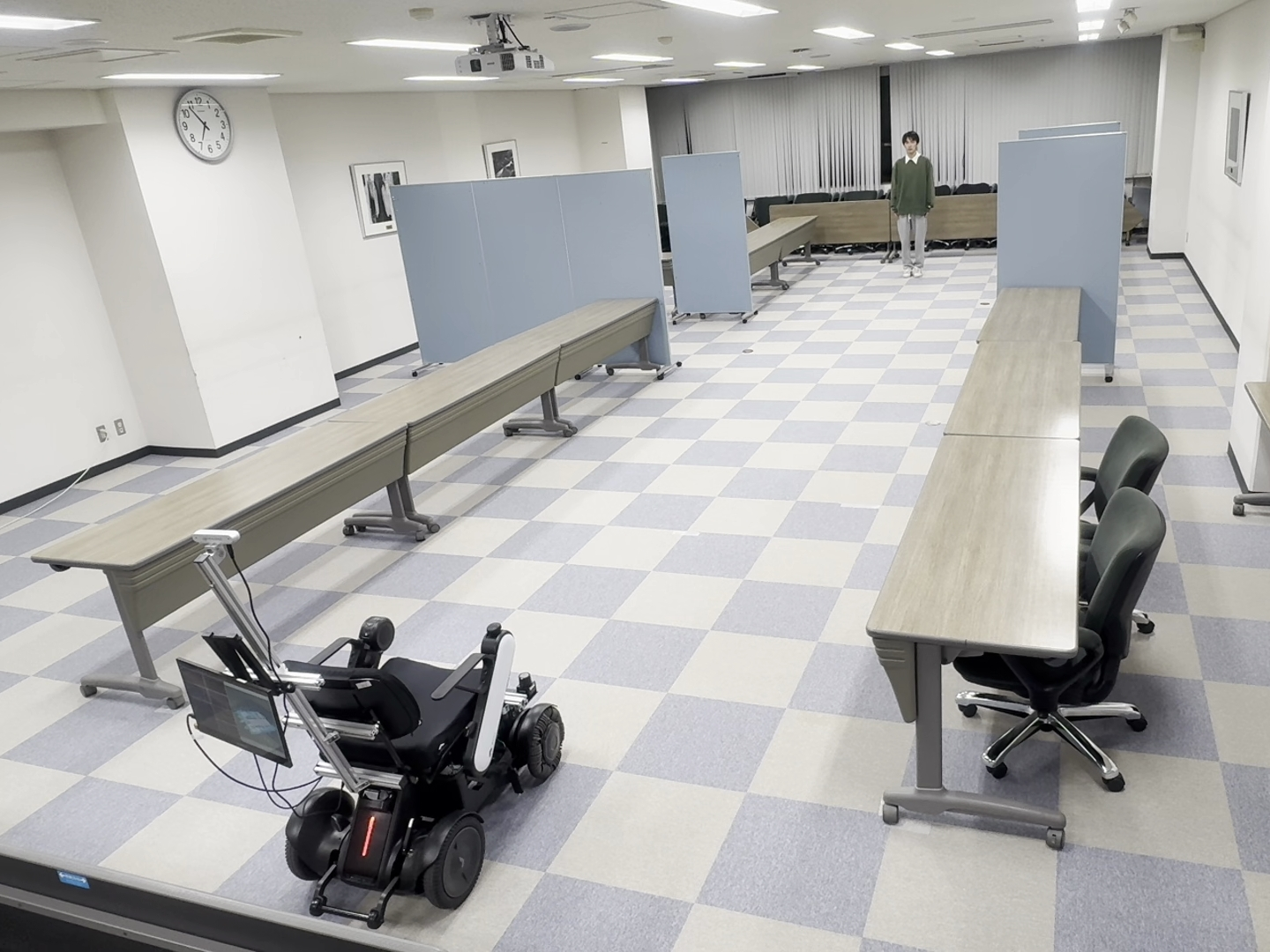}
    \caption{Experimental setup in an indoor intersection (main corridor: 3.0\,m $\times$ 10.0\,m/8.0\,m; cross corridor: 2.15\,m $\times$ 7.2\,m), 12 participants walk toward the robot.}
    \label{fig:experiment_setup}
\end{figure}

\subsection{Trajectory Prediction}

Trajectory prediction uses pre-trained HST models~\cite{salzmann2023robots}, which take the past 6
steps (2 seconds) of pedestrian positions and 3D skeletal keypoints as input and output distributions over $F = 12$
future steps (4 seconds) with $K = 6$ modes. The predicted distribution is modeled as a Gaussian Mixture Model (GMM):

\begin{equation} \label{eq:traj_gmm}
    \mathcal{P}(\tau) = \sum_{k=1}^{K} w_k
    \mathcal{N}(\tau; \mu_k, \Sigma_k),
\end{equation}

\noindent where $\tau$ denotes the future trajectory sequence over $F$ timesteps, $w_k$ is the mixture weight
of the $k$-th mode satisfying $\sum_{k=1}^{K} w_k = 1$, and $\mathcal{N}(\tau; \mu_k, \Sigma_k)$ is a Gaussian
distribution with mean trajectory $\mu_k \in \mathbb{R}^{F \times 2}$ and covariance matrix $\Sigma_k$.
For motion planning, the most probable mode $\hat{k}$ is selected:

\begin{equation} \label{eq:mode_select}
    \hat{k} = {\arg\max}_k\, w_k
\end{equation}

\noindent The mean trajectory $\mu_{\hat{k}}$ of this mode is passed to the cost-map generation module as the
representative predicted path. HST models are pre-trained on the JRDB dataset~\cite{martin2021jrdb} and predictions
are updated at approximately 3\,Hz in map coordinate systems.

\subsection{Adaptive Cost Map Generation}

The cost map follows a hierarchical layered architecture~\cite{lu2014layered} extended with a custom
layer for attention-aware planning. The three standard layers are: a \textit{Static Layer} encoding fixed obstacles from a pre-built map; an \textit{Obstacle Layer} representing dynamic obstacles from LiDAR; and an \textit{Inflation Layer} expanding obstacle boundaries by a fixed margin. To these, we add a \textit{Predicted Trajectory Layer} that directly
incorporates pedestrian trajectory predictions from HST as navigation costs, enabling the planner to account for
not only the pedestrian's current position but their anticipated future path.

The required safety margin around each predicted trajectory point is determined by whether the pedestrian can actively
participate in collision avoidance. Distracted pedestrians have limited situational awareness and cannot be expected
to adjust their path in response to an approaching robot; the robot must therefore maintain a larger clearance.
Attentive pedestrians, by contrast, can perceive the robot and respond accordingly, permitting a narrower margin.
Following Hall's proxemics framework~\cite{hall1966hidden}, personal distance ($>$0.45\,m) is adopted as the reference
for distracted pedestrians, with a smaller value applied for attentive pedestrians. The cost region radius $r$ is
defined as:

\begin{equation} \label{eq:cost_radius}
    r = \begin{cases}
        0.3\,\text{m} & \text{if } D(t) = 0
        \text{ (Attentive)}  \\
        0.5\,\text{m} & \text{if } D(t) = 1
        \text{ (Distracted)}
    \end{cases}
\end{equation}

Robot navigation is executed under ROS, with self-localization via Adaptive Monte Carlo Localization (AMCL)~\cite{dellaert1999monte}, global path planning via
A*~\cite{hart1968formal}, and local motion planning via TEB~\cite{rosmann2012trajectory}.

\begin{table}[tb]
    \centering
    \caption{Confusion matrix (left) and classification performance metrics (right) using Random Forest}
    \label{tab:confusion_and_metrics}
    % 左側：混同行列
    \begin{minipage}{0.55\linewidth}
        \centering
        \begin{tabular}{lcc}
            \toprule
                       & \multicolumn{2}{c}{Predicted}               \\
            \cmidrule(lr){2-3}
            Actual     & Attentive                     & Distracted  \\
            \midrule
            Attentive  & \textbf{35}                   & 2           \\
            Distracted & 1                             & \textbf{23} \\
            \bottomrule
        \end{tabular}
    \end{minipage}
    \hfill
    % 右側：評価指標
    \begin{minipage}{0.40\linewidth}
        \centering
        \begin{tabular}{lc}
            \toprule
            Metric    & Value (\%) \\
            \midrule
            Accuracy  & 95.1       \\
            Precision & 97.2       \\
            Recall    & 94.6       \\
            F1-Score  & 95.9       \\
            \bottomrule
        \end{tabular}
    \end{minipage}
\end{table}

\section{Experiment}

To evaluate the proposed framework, we conducted a within-subjects passing experiment in a T-shaped indoor
corridor environment (main corridor: 3.0\,m wide, 10.0\,m long in Scenario~1 and 8.0\,m long in
Scenario~2; cross corridor: 2.15\,m $\times$ 7.2\,m), as shown in Fig.~\ref{fig:experiment_setup}. 
All participants provided written informed consent prior to the experiment. 
Twelve participants (graduate students, aged 22--28) each completed 21 trials (7 per method), yielding 252 trials in total. 
The order was counterbalanced across participants using a Latin square design to mitigate order effects.

Each trial involved a single pedestrian walking toward the robot from the opposite end of the corridor. To
evaluate robustness across different attention transitions, four pedestrian behavioral states were
defined: \textbf{N} (normal walking throughout), \textbf{T} (texting throughout), \textbf{TN} (texting
then switching to normal mid-trial), and \textbf{NT} (normal then switching to texting mid-trial). These
states were distributed across two scenarios to assess the system's adaptability to both static and
dynamically changing attention states.

Three navigation methods were compared:
\begin{itemize}
    \item \textbf{Method~A (Baseline):} LiDAR-based obstacle-layer navigation with no pedestrian state
    estimation or trajectory prediction.
    \item \textbf{Method~B (Prediction):} Trajectory prediction incorporated into the cost map, but with
    a fixed cost region radius of $r = 0.5$\,m
    regardless of attention state. This radius is based on the same Hall proxemics-based personal-distance\cite{hall1966hidden} reference as
    the distracted-pedestrian setting in Method~C, making the comparison conservative.
    \item \textbf{Method~C (Ours):} Proposed method. Trajectory prediction combined with dynamic cost region radius adjustment ($r = 0.3$\,m or
    $0.5$\,m) based on attention classification according to Eq.~(\ref{eq:cost_radius}).
\end{itemize}

Evaluation combined subjective and quantitative measures. For subjective evaluation, participants
completed a 7-item questionnaire (Q1--Q7) on a 7-point Likert scale after each method, covering Safety~(Q1),
Comfort~(Q2), Smoothness~(Q3), Naturalness~(Q4), Predictability~(Q5), Trust~(Q6), and
Efficiency~(Q7). Three direct comparison questions (D1, D3, D4) asked participants to identify the
preferred method. Quantitative metrics included jerk RMS, minimum separation distance, maximum lateral
deviation, and travel time. Statistical analysis used the Kruskal--Wallis test with Dunn's post-hoc
correction for subjective ratings, and sign tests for pairwise preference comparisons.

For the attention classifier, a custom dataset was constructed by manually labeling 304 images drawn from
the public JRDB dataset~\cite{martin2021jrdb}. The dataset was split into 243 training images (146 distracted, 97 attentive) and 61 test images (37 distracted, 24 attentive). 

The platform was a WHILL Model CR wheelchair with Hokuyo UST-30LX LiDARs and an Intel RealSense D455 RGB-D camera ($1280\times720$, 15\,fps, mounted at 1.5\,m), with all computation on an NVIDIA Jetson AGX Orin running ROS Noetic.

\begin{figure}[tb]
    \centering
    \includegraphics[width=\linewidth]{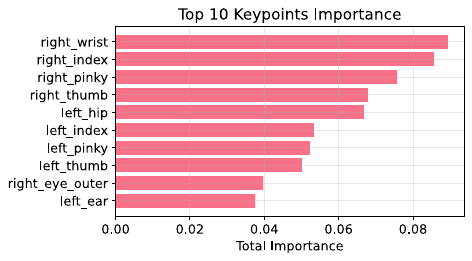}
    \caption{Feature importance of the Random Forest model. Feature importance scores indicating that wrist and shoulder positions are dominant factors.}
    \label{fig:classification_result}
\end{figure}

\begin{table}[tb]
    \caption{Quantitative evaluation results across three methods with mean $\pm$ SD.}
    \label{tab:quantitative}
    \centering
    \begin{tabular}{lccc}
        \toprule
        Metric & A & B & C (Ours)\\
        \midrule
        Jerk RMS (m/s$^3$) & 37.45$\pm$17.46 & \textbf{11.01$\pm$8.62} & 11.42$\pm$9.19 \\
            Min separation (m) & 1.74$\pm$0.68 & \textbf{0.94$\pm$0.28} & 1.01$\pm$0.31 \\
            Max lat.\ dev.\ (m) & 0.30$\pm$0.21 & 0.31$\pm$0.19 & \textbf{0.36$\pm$0.23} \\
            Travel time (s) & 16.8$\pm$2.1 & 17.0$\pm$1.9 & 16.9$\pm$2.0 \\
        \bottomrule
        \multicolumn{4}{l}{\footnotesize $^{*}$Method~A's 39\% detection rate is inflated; see Section~V-B.}
    \end{tabular}
\end{table}

\section{Results}

\subsection{Classifier Performance}
The Random Forest classifier achieved an accuracy of 95.1\%, precision of 97.2\%, recall of 94.6\%, and
F1-score of 95.9\% on the held-out test set (Table~\ref{tab:confusion_and_metrics}). Beyond
overall accuracy, the physical validity of the learned features provides evidence against overfitting to
environmental factors such as background or clothing. As shown in Fig.~\ref{fig:classification_result}, the
model assigns the highest importance to \textit{right\_wrist}, \textit{right\_elbow}, and
\textit{neck} --- precisely the keypoints activated by raising a forearm to hold a device and tilting the
head downward to view a screen. Lower-body keypoints such as ankles and knees contribute comparatively
little, indicating that the classifier responds to distraction-specific posture rather than general gait
characteristics.

\subsection{Quantitative Evaluation}

\subsubsection{Jerk RMS}
Both Methods~B and C produced significantly smoother robot trajectories than Baseline~A
(Table~\ref{tab:quantitative}). Mean jerk RMS values were $37.45 \pm 17.46$\,m/s$^3$ (A), $11.01 \pm
8.62$\,m/s$^3$ (B), and $11.42 \pm 9.19$\,m/s$^3$ (C). The reduction from A to the prediction-based
methods was highly significant ($p < 0.001$, Cohen's $d \approx 1.9$), while B and C showed no significant difference. Trajectory prediction is therefore the primary driver of objective motion smoothness --- attention-state classification does not alter this metric.

\subsubsection{Minimum separation distance}
Methods~B and C achieved mean separation distances of $0.94 \pm 0.28$\,m and $1.01 \pm 0.31$\,m
respectively, reflecting near-complete pedestrian detection via camera-based tracking. Method~A's value of $1.74 \pm 0.68$\,m is an artifact of its 39\% pedestrian detection
rate: when the obstacle layer fails to detect the pedestrian, the robot does not deviate, producing
spuriously large separation values in undetected trials. This metric is therefore not directly
comparable across all three methods.

\subsection{Subjective Evaluation}
Method~C achieved the highest mean scores across all seven subjective dimensions
(Table~\ref{tab:subjective}). The central finding, however, is the contrast between objective and
subjective outcomes across the three methods. Despite Method~B reducing jerk RMS by nearly three-fold
relative to Baseline~A, no significant difference between A and B was observed on any subjective
dimension ($p > 0.05$) --- participants perceived no improvement in safety, comfort, smoothness, or
predictability from trajectory prediction alone. Method~C broke this pattern decisively. Q3 Smoothness
showed the strongest effect: mean scores were 3.96~(A), 4.14~(B), and 4.89~(C) ($p = 0.003$, $\eta^2 = 0.29$,
large effect), with post-hoc tests confirming significant pairwise differences for both A vs.~C
($p_{\mathrm{adj}} = 0.004$) and B vs.~C ($p_{\mathrm{adj}} = 0.030$). Q5 Predictability
followed the same pattern: scores of 3.69~(A), 4.01~(B), and 4.48~(C) ($p = 0.034$, $\eta^2 = 0.14$,
large effect). These results suggest that attention-state-aware cost adjustment --- not
trajectory prediction alone --- is a key condition for improving the perceived quality of robot
navigation.

\begin{table}[tb]
    \caption{Subjective evaluation results (Q1--Q7). Mean scores are shown for each method.}
    \label{tab:subjective}
    \centering
    \begin{tabular}{lccc}
        \toprule
        Item & A & B & C (Ours)\\
        \midrule
        Q1 Safety        & 4.60 & 4.69 & \textbf{5.07} \\
        Q2 Comfort       & 4.24 & 4.42 & \textbf{4.82} \\
        Q3 Smoothness    & 3.96 & 4.14 & \textbf{4.89} \\
        Q4 Naturalness   & 4.13 & 4.22 & \textbf{4.67} \\
        Q5 Predictability& 3.69 & 4.01 & \textbf{4.48} \\
        Q6 Trust         & 4.49 & 4.64 & \textbf{4.96} \\
        Q7 Efficiency    & 4.38 & 4.57 & \textbf{4.93} \\
        \bottomrule
    \end{tabular}
\end{table}

In the direct comparison questions, Method~C received the most votes in D1 (``which method felt safest to
pass,'' 6/12; $\chi^2 = 3.50$, $p = 0.174$), though this did not reach statistical significance.
Participants who preferred Method~C in open-ended responses (D2) consistently cited earlier avoidance
initiation and smoother lateral displacement as primary reasons. In D3 (``which method was most
noticeable while using a smartphone''), 8 out of 12 participants selected Method~C ($\chi^2 = 6.00$,
$p = 0.050$), indicating that the robot's adaptive behavior was perceptible to the majority of
participants even while they were actively distracted.
In TN trials, participants reported that Method~C made the robot's avoidance direction clear after an initial face-to-face situation; in NT trials, comments mentioned unsteady or unclear avoidance direction. 
These responses suggest that attention-aware cost tuning can make some dynamic transitions more legible, but that instantaneous (frame-level) attention updates do not fully remove transient ambiguity during attention changes.

\subsubsection{Maximum lateral deviation}
Method~C produced consistently larger lateral path deviations than Method~B: 10 out of 12 participants
recorded greater maximum lateral deviation under C (sign test, $p = 0.039$, rank-biserial $r = 0.67$).
Although the mean difference was modest ($0.31$ vs.~$0.36$\,m), the strong directional consistency
across participants and the convergence with the D3 perceptual result together suggest that the wider
cost region under Method~C produced reliably greater lateral displacement --- displacement that was both
measurable and noticeable to pedestrians.

\subsubsection{Travel time}
No significant difference in travel time was observed across the three methods ($16.8 \pm 2.1$\,s,
$17.0 \pm 1.9$\,s, $16.9 \pm 2.0$\,s for A, B, and C respectively), confirming that the safety and
perceptual improvements of Method~C carry no cost to navigation efficiency.

\section{Discussion}

Methods~B and C are objectively indistinguishable in motion smoothness --- their jerk RMS values differ by
less than 4\% --- yet Method~C scores significantly higher on every subjective dimension, with large
effect sizes on both Smoothness ($\eta^2 = 0.29$) and Predictability ($\eta^2 = 0.14$). This divergence
between objective and subjective outcomes points to a specific mechanism: what participants responded to was
not how smoothly the robot moved, but how consistently and directionally it moved away from them. Method~C
always widens its path relative to a distracted pedestrian in the same lateral direction, producing
avoidance behavior that is legible and anticipatable from the pedestrian's perspective. Method~B, with its
fixed cost radius, generates avoidance patterns that vary with pedestrian proximity, making the robot's
intent harder to read. The lateral deviation data supports this directly --- 10 out of 12 participants
recorded larger maximum lateral deviation under C ($r = 0.67$), and 8 out of 12 recognized this
behavioral difference even while actively distracted (D3, $p = 0.050$). Together these results form a
consistent picture: attention-aware cost adjustment can produce navigation behavior that is more legible
to the people around it.

The A vs.\ B result warrants separate consideration as a finding in its own right. Method~B reduced jerk
RMS by nearly three-fold relative to Baseline~A, yet participants perceived no improvement across any of
the seven subjective dimensions ($p > 0.05$). This decoupling has a direct design implication: objective motion quality alone may be insufficient unless avoidance behavior is legible to pedestrians.

This study has several limitations. First, the participant pool of 12 graduate students aged 22--28 limits
generalizability; validation with broader populations, including older adults and non-technical users, is an
important next step given the service robot deployment contexts this work targets.
Second, the current attention classifier operates instantaneously at the frame level, so it does not explicitly model temporal transitions in attention or recovery after distraction; incorporating temporal attention dynamics is an important extension of the present framework.
The binary classification also simplifies a continuous phenomenon, and replacing the threshold in Eq.~(\ref{eq:cost_radius}) with a soft cost weight derived from the classifier's posterior probability would allow the safety margin to scale gradually with estimated distraction level.
Third, the forward-facing RGB-D camera restricts pedestrian tracking to the frontal cone; a 360-degree perception system would enable the robot to maintain awareness after a pedestrian has passed.
Finally, the current evaluation considers only single-pedestrian scenarios --- real deployments involve multiple simultaneous pedestrians with potentially different attention states, requiring priority-based cost allocation across individuals.
Each represents a concrete and tractable extension of the present framework.

\section{Conclusion}

This paper presented an attention-aware navigation framework that classifies pedestrians as attentive or distracted in real-time from 3D skeletal features and dynamically adjusts cost-map safety margins for proactive avoidance. A within-subjects user study with 12 participants across 252 trials suggests that trajectory prediction alone may be insufficient: subjective improvements in smoothness ($p = 0.003$) and predictability ($p = 0.034$) emerged only when attention-state classification was introduced, with no increase in travel time. These results indicate that perceived navigation quality depends not on motion smoothness per se, but on whether avoidance behavior is legible to pedestrians. Future work will address temporal attention dynamics, continuous distraction modeling, multi-pedestrian priority allocation, and broader population validation.

%%%%%%%%%%%%%%%%%%%%%%%%%%%%%%%%%%%%%%%%%%%%%%%%%%%%%%%%%%%%%%%%%%%%%%%%%%%%%%%%

\bibliographystyle{IEEEtran}
\bibliography{bibfile}

\vfill

\end{document}